\documentclass{article}

\usepackage{inputenc}
\newcommand{\cut}[1]{}
\usepackage{graphicx}
\usepackage{booktabs}
\usepackage{color}
\usepackage{amsmath}
\usepackage{amsthm}
\usepackage{amssymb}

\usepackage{natbib}
\usepackage{url}
\usepackage{wrapfig}
\usepackage{xcolor} 
\usepackage[noend, boxed, linesnumbered, algo2e]{algorithm2e}
\usepackage{algpseudocode}
\usepackage{algorithm}
\usepackage{algorithmicx}
\usepackage{algcompatible}
\usepackage{epstopdf}
\usepackage{multirow}
\usepackage{rotating}
\usepackage{relsize}
\usepackage{appendix}
\usepackage{bbm}
\usepackage{mathtools}

\usepackage{tabularx}
\usepackage{graphicx}
\usepackage{adjustbox}

\newtheorem{definition}{Definition}[section]

\usepackage{appendix}
\usepackage{amsthm,amssymb}
\usepackage{tcolorbox}

\usepackage{arxiv}

\usepackage{hyperref}       
\usepackage{url}            
\usepackage{booktabs}       
\usepackage{amsfonts}       
\usepackage{nicefrac}       
\usepackage{microtype}      
\usepackage{lipsum}		
\usepackage{graphicx}
\usepackage{natbib}
\usepackage{doi}

\title{Detecting Important Patterns Using Conceptual Relevance Interestingness Measure}

\author{{Mohamed-Hamza Ibrahim} \\
	\cut{Department of Computer Science and Engineering, \\ }Universit\'{e} du Qu\'{e}bec en Outaouais, Canada \\
	\texttt{mohamed.ibrahim@polymtl.ca} \\
	\And
	{Rokia Missaoui} \\
	\cut{Department of Computer Science and Engineering, \\ }Universit\'{e} du Qu\'{e}bec en Outaouais, Canada \\
	\texttt{Rokia.missaoui@uqo.ca}
	\AND
	{Jean Vaillancourt} \\
	\cut{Department of Decision Sciences,  \\ }HEC Montreal, Canada \\
	\texttt{jean.vaillancourt@hec.ca}
}

\renewcommand{\shorttitle}{}

\begin{document}
\maketitle

\begin{abstract}
Discovering meaningful conceptual structures is a substantial task in data mining and knowledge discovery applications. While off-the-shelf interestingness indices defined in Formal Concept Analysis may provide an effective relevance evaluation in several situations, they frequently give inadequate results when faced with massive formal contexts (and concept lattices), and in the presence of irrelevant concepts. In this paper, we introduce the \textit{Conceptual Relevance} ($\mathcal{CR}$) score, a new scalable interestingness measurement for the identification of actionable concepts. From a conceptual perspective, the minimal generators provide key information about their associated concept intent. Furthermore, the relevant attributes of a concept are those that maintain the satisfaction of its closure condition. Thus, the guiding idea of $\mathcal{CR}$ exploits the fact that minimal generators and relevant attributes can be efficiently used to assess concept relevance. As such, the $\mathcal{CR}$ index quantifies both the amount of conceptually relevant attributes and the number of the minimal generators per concept intent. Our experiments on synthetic and real-world datasets show the efficiency of this measure over the well-known stability index.
\end{abstract}

\keywords{Formal Concept Analysis \and Pattern selection \and Interestingness index.}

\section{Introduction}
A wide range of crucial problems in different sub-disciplines including data mining, knowledge discovery, social network analysis, bioinformatics, and machine learning can be formulated as a pattern mining task. Inspired by the  mathematical power of Formal concept Analysis (FCA) \citep{Ganter+1999}, the lattice formalisation is always rich with substantial local conceptual structures that are important to data mining tasks \citep{belohlavek2011selecting}. In the lattice, each element captures a formal concept and the whole lattice represents a hierarchy of concepts. Unfortunately, it could contain a large amount of irrelevant local structures, which traditionally impose a discrepancy between a real and current settings of a given formal context. The irrelevant objects emerge due to a number of reasons, such as the imprecise and inaccurate collection of the data, e.g., erroneously inserting or omitting objects when describing some attributes. On the flip side of the coin, the irrelevant attributes are redundant and frequently appear due to the completeness property of the lattice. They potentially lead to a high complexity even for small datasets \citep{belohlavek2011selecting}. In general, whether it is an irrelevant attribute or object inside a formal concept, it does not sufficiently contribute to the actionability of such a pattern. That is, the irrelevant element often provides useless information when it is involved in patterns extracted from the formal context. So, its removal from the concept has no impact on its conceptual structure and significantly purifies domain description and semantics. Traditionally, there are three main strategies to mine interesting patterns in a concept lattice \citep{Kuznetsov2018}:
 (i) formal context reduction using methods such as Singular Value Decomposition or Non-Negative Matrix Decomposition, (ii) background knowledge (e.g., taxonomy or weight on attributes) or constraint consideration, and (iii) concept selection. In this paper we will focus on the third topic which involves picking out a subset of concepts based on interestingness measures \citep{Belohlavek2013}. In the FCA literature, several selection measures (see \citep{Kuznetsov2018} for a detailed survey) have been proposed such as concept probability\cut{\citep{Kuznetsov2018}}, robustness\cut{\citep{tatti2014finding}}, separation\cut{\citep{klimushkin2010approaches}}, Monocle \citep{torim2008sorting}, $\delta$-tolerance closed frequent itemsets \citep{cheng2006delta}, margin-closed itemset \citep{moerchen2011efficient} and predictability \citep{belohlavek2013basic} among others. On the basis of the Galois connection between the set of objects and the set of attributes, the closure and derivation conditions of formal concepts are often decisive properties for measuring the importance of these patterns. As such, the stability  index \citep{kuznetsov2007reducing}, which depends primarily on these two properties, has recently been introduced as the most prominent index for assessing the concept quality \citep{Kuznetsov2018}. 

Although the stability index serves as a good indication of how much relevant detail is expected inside the concept, it is known that computing stability is $\#$P-complete \citep{roth2008succinct}, which often requires an exponential time complexity in the size of the intent (or extent), i.e., $O(2^{|\text{Intent}|} \times |\mathcal{G}| \times |\mathcal{M}|)$. For example, we could need to perform more than $2^{22}\approx 4.1$ million computational comparisons to calculate the stability of a concept with an intent size equal to $22$. In \citep{roth2008succinct}, it was demonstrated that the stability can be computed based on the generators of the concept intent. However, \cut{finding the set of generators associated with a concept intent requires the enumeration through the concept upper covers (i.e., immediate successors). }this often has a time and space complexity of $O(|\mathcal{G}|^2 \cdot |\mathcal{M}| \cdot L^2)$ which is at least quadratic in the size of the input lattice $L$ \citep{zhi2014calculation,roth2008succinct}. This is problematic even for small concepts since the lattice size could be exponential with respect to the size of the context. Furthermore, stability is dependent on generators but not necessarily minimal ones, and it is widely known that using generators causes an overestimation of concept quality, resulting in redundant association rules, including implications. 

In this paper, to avoid the limitations of stability, we introduce the \textit{Conceptual Relevance} ($\mathcal{CR}$) index. At the conceptual level, our overall approach to the $\mathcal{CR}$ index consists of the following basic elements. First, we exploit the fact that using minimal generators of a concept intent only frequently results in less redundant (or more important) patterns (e.g., implications and association rules). Second, we formulate what is meant by relevant attributes. More precisely, these attributes are often essential for maintaining stable conceptual structures. As a result, we design the Conceptual Relevance of the concept to concurrently quantify the amounts of minimal generators and relevance attributes in its intent, packed in one measure. The computation of the $\mathcal{CR}$ index requires a polynomial time in the size of the concept's upper covers, and is therefore quite fast in practice. 

The rest of the paper is organized in the following manner. Section~\ref{Back} recalls some basic definitions of FCA and stability index. Section~\ref{RainbowSect} explains our proposed Conceptual Relevance index for measuring the concept relevancy in further more detail. In Section~\ref{Exp} we conduct a thorough experimental study with a detailed discussion. Finally, Section~\ref{Con} presents our conclusions.

\section{Background}\label{Back}
\begin{figure}[!htbp]
  \begin{center}
  \includegraphics[width=2.3in,height=1.6in]{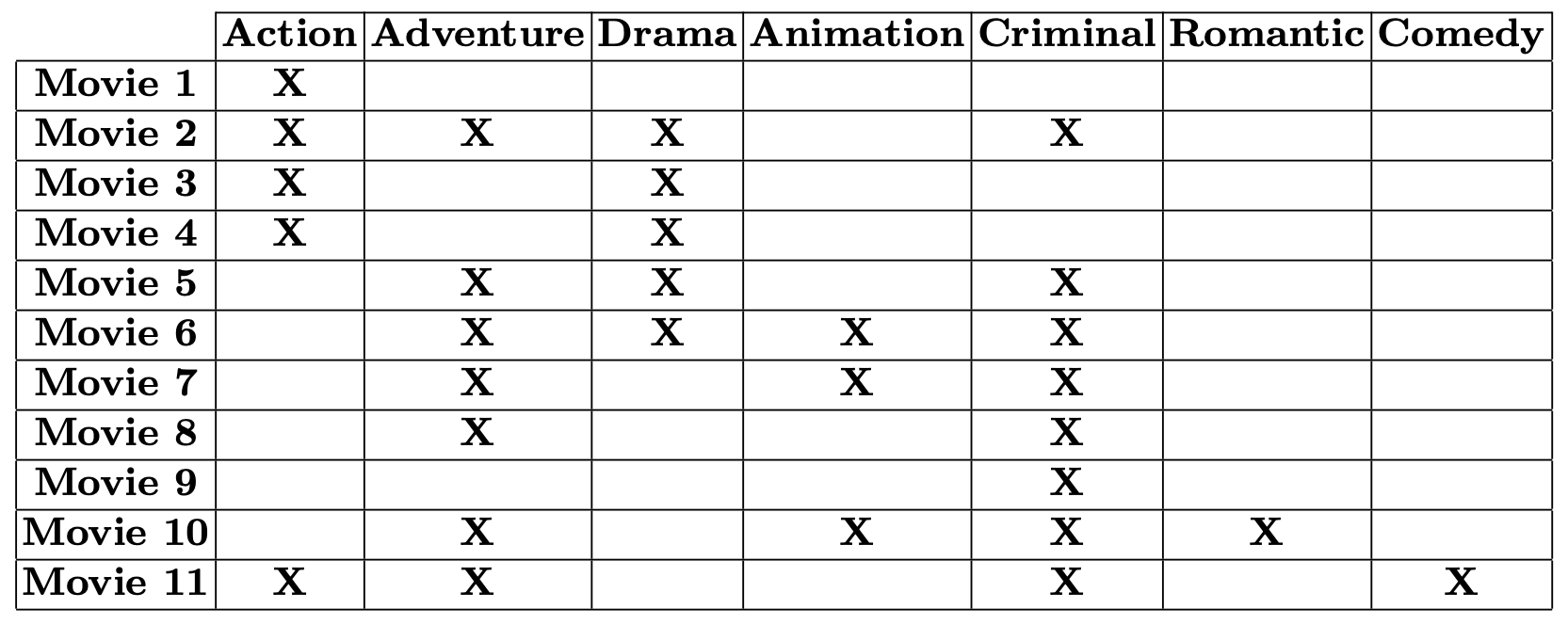}
  \includegraphics[width=3.3in,height=3.2in]{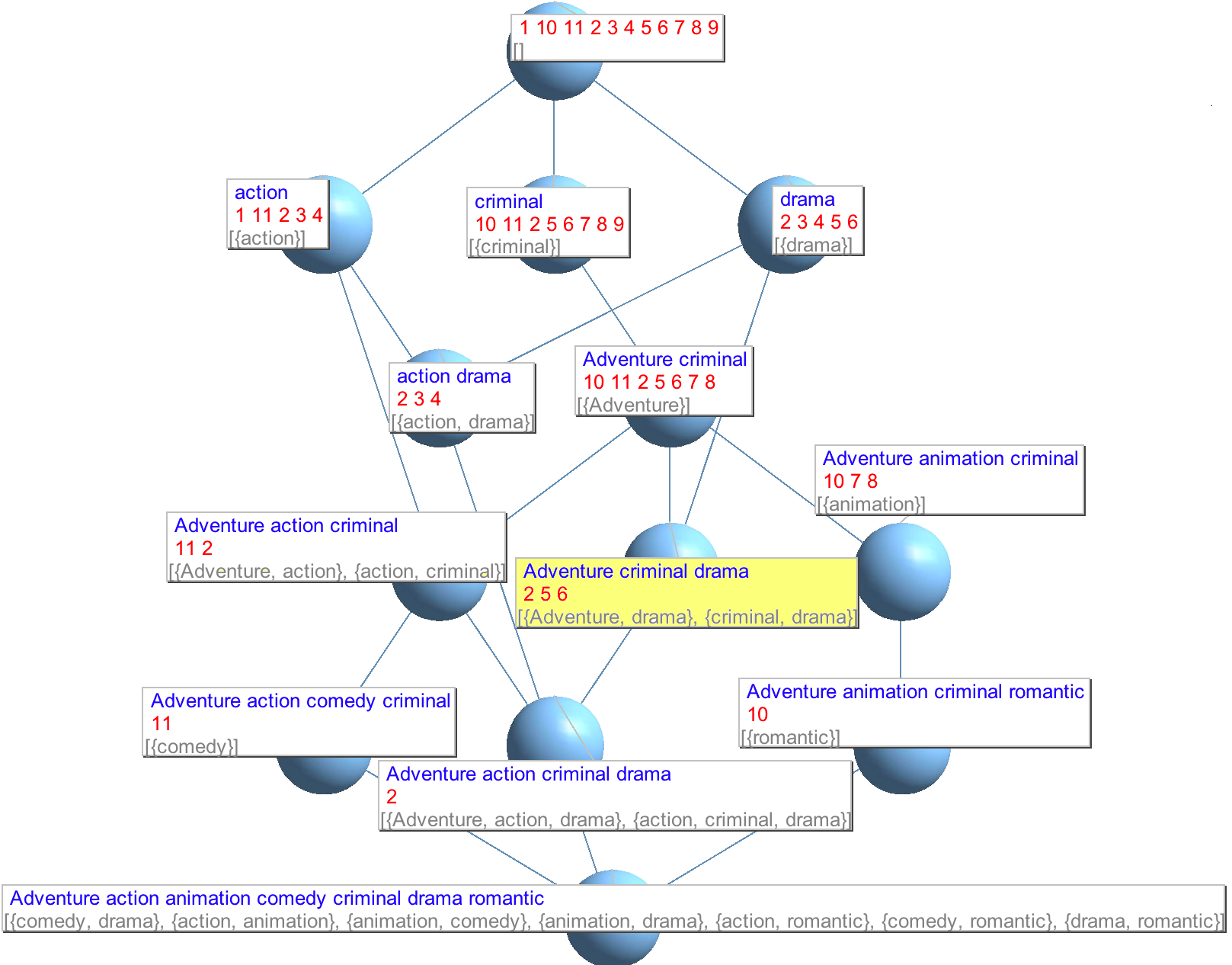}
  \end{center}
\caption{(left) The \textit{Movies} formal context and (right) its corresponding concept lattice.}
\label{fig1}
\end{figure}

To illustrate the basic notions in Formal Concept Analysis (FCA) we use the \emph{Movies} formal context with 11 objects and 7 attributes (see Figure~\ref{fig1}).

\subsection{Formal Concept Analysis} 
Formal concept analysis (FCA) is a general mathematical framework for building clusters defined as object sets sharing common attributes. Here, we briefly recall the FCA terminology \citep{Ganter+1999}. The starting point of an FCA analysis is a formal context, and the main output is a concept lattice (see Figure~\ref{fig1}).
\begin{definition} [Formal context]
A \textit{formal
context} is a triple $\mathbb{K} = (\mathcal{G},\mathcal{M},\mathcal{I})$, where $\mathcal{G}$ is the set of objects, $\mathcal{M}$ is the set of attributes, and $\mathcal{I}$ is a binary relation between $\mathcal{G}$ and $\mathcal{M}$ with $\mathcal{I} \subseteq \mathcal{G} \times \mathcal{M}$. For $g \in \mathcal{G}, m \in \mathcal{M}, (g,m) \in \mathcal{I}$ holds iff the object $g$
has the attribute $m$. 
\end{definition}
Given arbitrary subsets $A \subseteq \mathcal{G}$ and $B \subseteq \mathcal{M}$, the Galois connection can be defined by the following derivation operators: $A^{\prime{}} = \{m \in \mathcal{M} \mid \forall g \in A, (g,m) \in \mathcal{I} \}$ and $B^{\prime{}} = \{g \in \mathcal{G} \mid \forall m \in B, (g,m) \in \mathcal{I} \}$,
where $A^{\prime{}}$ is the set of attributes common to all objects of $A$ and $B^{\prime{}}$ is the set of objects sharing all attributes from $B$. The closure operator $(·)^{\prime{}\prime{}}$ implies the double application of $(·)^{\prime{}}$.

\begin{definition} [Formal concept]
The pair $c=(A,B)$ is called a \textit{formal concept} of $\mathbb{K}$ with \textit{extent} $A$ and \textit{intent} $B$ iff $A^{\prime{}\prime{}}=A$ and $B^{\prime{}\prime{}}=B$. 
\end{definition}

For example, in Figure~\ref{fig1}, $\tilde{c}=(\{2,5,6\},\{\text{adventure,criminal,drama}\})$ is a formal concept with extent $A=\{2,5,6\}$ and intent $B=\{\text{adventure,criminal,drama}\}$. In the sequel, we use $\tilde{c}$ as our illustrative example to support the understanding of definitions and principles related to the Conceptual relevance index.     

A partial order exists between two concepts $c_1=(A_1,B_1)$ $\preceq$ and $c_2=(A_2,B_2)$ if $A_1 \subseteq A_2 \iff  B_1 \supseteq B_2$.    
The set $\mathcal{C}$ of all concepts together with the partial order form a concept lattice.

\begin{definition}[Covers and faces]\label{LUc}
For two formal concepts $c_1=(A_1,B_1) \preceq c_2=(A_2,B_2)$, if $\nexists \; c3=(A_3,B_3) \text{ s.t. } \cut{(A_1,B_1) \preceq  (A_3,B_3) \preceq (A_2,B_2)  } A_1 \subseteq A_3 \subseteq A_2$,
\cut{\begin{equation}
\cut{(A_1,B_1) \preceq (A_2,B_2), }\nexists \; c3=(A_3,B_3) \text{  such that } \cut{(A_1,B_1) \preceq  (A_3,B_3) \preceq (A_2,B_2)  } A_1 \subseteq A_3 \subseteq A_2 \iff  B_1 \supseteq B_3  \supseteq B_2 
\end{equation}
or 
\begin{equation}
A_1 \subseteq A_3 \subseteq A_2 \iff  B_1 \supseteq B_3  \supseteq B_2
\end{equation}}
then $c_1$ is a \textit{lower cover} of $c_2$, and $c_2$ is an \textit{upper cover} of $c_1$, denoted by $c_1 \prec c_2$ and  $c_2 \succ c_1$ respectively. The \textit{intentional face} $f_{\text{int}}(c,c_u)$ of a concept $c=(A,B)$  w.r.t. its $u$-th upper cover concept, $c_u=(A_u,B_u) \in \mathcal{U}(c)$, is the difference between their intent sets \citep{pfaltz2002closed}, \textit{i.e.},
$f_{\text{int}}(c,c_u) = B \setminus B_u$, where $\mathcal{U}(c)$ is the set of the upper covers of $c$.
\end{definition}

\cut{\begin{definition}[Concept `intentional' face]\label{intface} \citep{pfaltz2002closed,pfaltz2002scientific} 
The \textit{intentional face} $f_{\text{int}}(c,c_u)$ of a concept $c=(A,B)$  w.r.t. its $u$-th upper cover concept, $c_u=(A_u,B_u) \in \mathcal{U}(c)$, is the difference between their intent sets as:
$f_{\text{int}}(c,c_u) = B \setminus B_u$.
\end{definition}
\cut{The intentional face of $\tilde{c}$ w.r.t. its upper cover $c_u=(\{2,5,6,7,8,10,11\},\{\text{adventure,criminal}\})$ is $f_{\text{int}}(c,c_u) = \{\text{action}\}$}

\begin{definition}[Concept `extensional' face]\label{extface} 
The \textit{extensional face} $f_{\text{ext}}(c,c_l)$ of a concept $c=(A,B)$  w.r.t. its $l$-$th$ lower cover concept, $c_l=(A_l,B_l) \in \mathcal{L}(c)$, is the difference between their extent sets as:
$f_{\text{ext}}(c,c_l) = A \setminus A_l$.
\end{definition}
The extensional face of $\tilde{c}$ w.r.t. its lower cover $c_l=(\{2\},\{\text{action,adventure,criminal,drama}\})$ is $f_{\text{ext}}(c,c_l) = \{11\}$. We use $\mathcal{F}_{\text{ext}}(c)$ and $\mathcal{F}_{\text{int}}(c)$ to denote the extensional and intentional families of faces of a concept $c$ w.r.t. its lower and upper covers respectively.}

\begin{definition}[Generator \citep{bastide2000mining}]\label{ming} 
Given a concept $c=(A,B)$ in a formal context $\mathbb{K} = (\mathcal{G},\mathcal{M},\mathcal{I})$, a subset $h_i \subseteq B$ is called a \textit{generator} of the intent $B$ of $c$ iff $h_i^{\prime{}\prime{}}=B$, and it is a \textit{minimal generator} when $\nexists h_j \subseteq h_i$ such that $h_j^{\prime{}\prime{}} =B$. We use $\mathcal{H}_{c}$ to denote the set of minimal generators of the intent of concept $c$.
\end{definition}

\subsection{Concept Interestingness}
Interestingness measures of a formal concept $c=(A, B)$ are widely used to assess its \textit{relevancy}. In this context, the stability index $\sigma(c)$ of $c$ has been found to be prominent for selecting actionable concepts \citep{Kuznetsov2018}. 

\begin{definition}[Stability Index] 
\cut{\footnote{Note that since both intentional and extensional stability indices provide a dual measure of concept noise, then for notational simplicity, the subscripts $in$ and $ex$ are omitted, and we will generally use $\sigma(c)$ to mean the intentional stability in the rest of the paper.}} Let $\mathbb{K} = (\mathcal{G},\mathcal{M},\mathcal{I})$ be a formal context and $c=(A,B)$ a formal
concept of $\mathbb{K}$. The \textit{intentional stability} $\sigma_{\text{in}}(c)$ is: 
\begin{equation}\label{stability1}
    \sigma_{\text{in}}(c) = \frac{\mid \{e \in \mathcal{P}(A) | e^{\prime{}}= B\}\mid}{2^{|A|}}  = \frac{\mid \{e \in \mathcal{P}(A) | e^{\prime{}\prime{}}= A\}\mid}{2^{|A|}}
\end{equation}
\cut{In the dual way, the \textit{extensional stability} $\sigma_{\text{ex}}(c)$ is defined as follows:
\begin{equation}\label{stability2}
		\sigma_{\text{ex}}(c) = \frac{\mid \{e \in \mathcal{P}(B) | e^{\prime{}}= A\}\mid}{2^{|B|}}  = \frac{\mid \{e \in \mathcal{P}(B) | e^{\prime{}\prime{}}= B\}\mid}{2^{|B|}}
\end{equation}}
\end{definition}

where $\mathcal{P}(A)$ is the power set of $A$. 
In Equation~\eqref{stability1}, the intentional stability $\sigma_{\text{in}}(c)$ measures the strength of dependency between the intent $B$ and the objects of the extent $A$. More precisely, it expresses the probability to maintain $B$ closed when a subset of noisy objects in $A$ are deleted with equal probability. This measure quantifies the amount of noise that causes overfitting in the intent $B$. \cut{In a dual way, the extensional stability $\sigma_{\text{ex}}(c)$ (see Eq.~\eqref{stability2}) measures the portion of relevance in the extent $A$. For example, the intentional stability of the concept $c=(\{2,5,6\},\{\text{adventure,criminal,drama}\})$ is $\sigma_{\text{in}}(\tilde{c})=\frac{6}{2^3}=0.75$ whether its extensional stability is $\sigma_{\text{ex}}(\tilde{c})=\frac{2}{2^3}=0.375$. Incoherently, this concept $c$ is relevant based on its intentional stability and is irrelevant according to its extensional stability (cf. limitation A2).}     

\section{Conceptual Relevance Measure}\label{RainbowSect}
To set the stage for how the Conceptual Relevance quantifies the relevance of the formal concept, the first thing we do is to define\cut{ conceptually} relevant \cut{and irrelevant }attributes through the lens of FCA. 
\begin{definition}[Conceptually Relevant Attribute \citep{Ganter+1999,ganter2012formal}]\label{noistattr}
For a formal concept $c=(A,B)$, an attribute $m \in B$ is \textit{conceptually relevant} if\cut{ its removal from $B$ violates the derivation condition of the $c$ intent as: $(B\setminus\{m\})^{'} \neq A$.}
\begin{equation}\label{anc2}
(B\setminus\{m\})^{'} \neq A \cut{\text{\; and \;} \forall h \in \mathcal{H}_{c}, \; m \in h}.
\end{equation}
\end{definition}
That is, the concept $c$ does not preserve its local conceptual structure after removing $m$ from its intent (and accordingly from $\mathcal{M}$). From a statistical perspective, this means that the attribute $m$ has a significant statistical influence, and mainly depends on the distribution of the concept's intent parameter. Intuitively, this implies that the attribute $m$ contains certain relevant information in $c$. Thus, taking it off from the concept $c$ intent results in the loss of essential conceptual information (which clearly appears through the expansion of its extent). 
For instance, the attribute `drama’ is conceptually relevant in $\tilde{c}$ since its removal results in an expansion of the extent and a violation of the derivation condition of the intent, i.e., $(\{\text{adventure,criminal}\})^{'} = \{2,5,6,7,8,10,11\} \neq \{2,5,6\}$. 
On the contrary, `adventure’ is not a conceptually relevant attribute in $\tilde{c}$ since its removal from the intent does not violate the derivation condition, i.e., $(\{\text{criminal,drama}\})^{'} = \{2,5,6\}$.
At this point, we have paved the way for Conceptual Relevance index.

\begin{definition}[Conceptual Relevance Index $\mathcal{CR}(c)$]
The \textit{intentional Conceptual Relevance} of a concept $c=(A,B)$, in the formal context $\mathbb{K} = (\mathcal{G},\mathcal{M},\mathcal{I})$, can be computed as:
\begin{equation}\label{cr1}
 \mathcal{CR}_{\text{in}}(c) = \mathcal{F} \big(\alpha_{\text{in}}(c),\beta_{\text{in}}(c)\big)  
\end{equation}
where 
 \begin{equation}\label{cr2}
 \begin{array}{lll}
\alpha_{\text{in}}(c) =\begin{cases}
\frac{|\{m \in B | (B\setminus\{m\})^{'} \neq A \}|}{|B|} & \text{if } B \neq  \emptyset\\
0 & \text{Otherwise}.
 \end{cases}
 \end{array}
 \end{equation}
 and 
 \begin{equation}\label{cr3}
 \begin{array}{lll}
\beta_{\text{in}}(c) =\begin{cases}
\frac{|\mathcal{H}_{c}|}{2^{|B|}-2} & \text{if } |\mathcal{H}_{c}|>1 \And |B| >1 \\
0 & \text{Otherwise}.
 \end{cases}
 \end{array}
 \end{equation} 
\end{definition}

Algorithm~\ref{CRalgo} gives the pseudo-code for calculating the \textit{Conceptual Relevance} score in Eqs.~\ref{cr1}-\ref{cr3}. It takes as input the concept and its set of upper covers. At the first step, it computes the $\alpha_{\text{in}}$ term by iterating through the attributes of the intent $B$ to count the number of the conceptually relevant ones that satisfy the derivation condition as in Eq.~\ref{cr2} (Lines 2-10). From a conceptual perspective, the $\alpha_{\text{in}}$ term quantifies the ratio of the relevant attributes that exist in the concept intent $B$. For the $\beta_{\text{in}}$ term in Eq.~\eqref{cr3}, we calculate the set of minimal generators of $c$ as presented in \citep{szathmary2014fast} (Line 11). We then count the number of minimal generators out of all possible generators of intent $B$, as in Eq.~\ref{cr3} (Lines 12-14). Note that the goal behinds taking minus $2$ in the denominator of Eq.~\ref{cr3} is to exclude the trivial empty set and the whole intent. From a pattern mining perspective, the $\beta_{\text{in}}$ term quantifies the number of potential local relevant substructures inside the concept intent $B$. Note that due to the fact that in the formal context the attributes outside the scope of a given concept intent do not have any influence on its conceptual structure. Therefore, only the attributes of the concept intent should be used to normalize its $\mathcal{CR}(c)$ terms. As a result, the size of the intent $|B|$ and the one of its power set $2^{|B|}$ serve as normalization factors to scale both $\alpha_{\text{in}}$ and $\beta_{\text{in}}$ respectively. Finally, the algorithm computes a function $\mathcal{F}$ of both $\alpha_{\text{in}}$ and $\beta_{\text{in}}$ relevance terms. $\mathcal{F}$ can be any activation function applied to squash additive, multiplicative, divisor, logarithmic or other linear or non-linear relationships between the $\alpha_{\text{in}}$ and $\beta_{\text{in}}$ terms. For instance, $\mathcal{F}$ can simply be the arithmetic average that computes the linear additive relationship of the two terms. For example, the Conceptual Relevance score using the arithmetic average as an activation $\mathcal{F}$ of our concept $\tilde{c}$ is $\mathcal{CR}(\tilde{c})=\frac{1}{2}[\frac{1}{3}+\frac{2}{2^3-2}]=0.333$.

\begin{algorithm}[!htb]
\caption{Computing Conceptual Relevance Index.}
\KwIn{Concept $c=(A,B)$, set of upper covers $\mathcal{U}(c)$, activation function $\mathcal{F}$.}
\KwOut{Conceptual Relevance score $\mathcal{CR}(c)$.}
  \begin{algorithmic}[1] 
      \STATE $\alpha \leftarrow \beta \leftarrow 0$;
      \STATEx \texttt{// 1. Calculate $\alpha_{\text{in}}$ term.}
      \IF{$B \neq \emptyset$}
         \STATE $\text{Count } \leftarrow 0$;
         \FOR{\text{each attribute $m$ in $B$}}
            \IF{$(B\setminus\{m\})^{'} \neq A$}
                \STATE $\text{Count } \leftarrow \text{Count}+1$;
            \ENDIF
        \ENDFOR
      \ENDIF
         \STATE $\alpha \leftarrow \frac{\text{Count}}{|B|}$; 
    \STATEx \texttt{// 2. Calculate $\beta_{\text{in}}$ term.}
        \STATE $\mathcal{H}_{c} \leftarrow Minigen(c,\mathcal{U}(c))$; \texttt{// Calculate minimal generators.}
       \IF{$|B| > 1$ and $|\mathcal{H}_{c}| > 1$}
         \STATE $ \beta \leftarrow \frac{|\mathcal{H}_{c}|}{2^{|B|}-2}$;
        \ENDIF
       \STATE $\mathcal{CR}(c) \leftarrow \mathcal{F}(\alpha,\beta)$;
           \STATE $\text{\textbf{Return}} \; \mathcal{CR}(c)$;
    \end{algorithmic} \label{CRalgo}
\end{algorithm}

\paragraph{Complexity Analysis.} The calculation of the $\alpha_{\text{in}}$ term has time and space complexity of $O(|B| \times |\mathcal{G}| \times |\mathcal{M}|)$ since we store and proceed all attributes to check their relevancy condition. The $\beta_{\text{in}}$ term needs time and space complexity of $O(|\mathcal{U}(c)| \times |\mathcal{H}_{c}|)$ because we have to store and process all upper-covers of $c$ to calculate the faces, and then iteratively check their intersections with each element in the progressive minimal generator set. Because the time complexity  $O(|B| \times |\mathcal{G}| \times |\mathcal{M}|)$ of the first term subsumes the second one $O(|\mathcal{U}(c)| \times |\mathcal{H}_{c}|)$, the CR index has total time complexity of $O(|B| \times |\mathcal{G}| \times |\mathcal{M}|)$.

\section{Experimental Evaluation}\label{Exp}
The objective of our empirical evaluation is to address these two key questions:
\begin{itemize}
	\item (\textbf{Q1}.) Is the $\mathcal{CR}$ index empirically accurate compared to the state-of-the-art indices for assessing the relevance of formal concepts? \textit{We seek to empirically analyze the accuracy of the Conceptual Relevance index}.
	\item (\textbf{Q2}.) Is the $\mathcal{CR}$ index faster than the state-of-the-art interestingness indices? \textit{We want to validate the efficiency of the Conceptual Relevance index.}
\end{itemize}

\subsection{Methodology}
In order to obtain robust answers to Questions Q1 and Q2, we first select the following (synthetic$^\ddagger$ and real-life$^\star$) datasets\footnote{Publicly available: \url{https://github.com/tomhanika/conexp-clj/tree/dev/testing-data}; \url{http://icfca2012.markuskirchberg.net/index.php?page=webDataSets}}: 
\begin{itemize}
    \item $^\ddagger$\textbf{Dirichlet} \citep{felde2019formal} is a random formal context generated using the Dirichlet model generator\footnote{Publicly available: \url{https://github.com/maximilian-felde/formal-context-generator}}   
 \item  $^\ddagger$\textbf{CoinToss} \citep{felde2020null} is a random formal context generated by indirect Coin-Toss model generator
 \item  $^\star$\textbf{LinkedIn} \citep{nr} contains information about some computer science experts' job experiences/skills gleaned from their interconnected LinkedIN profiles
 \item $^\star$\textbf{Diagnosis} \citep{czerniak2003application} includes a set of symptoms to make a medical diagnosis of whether a patient has a bladder inflammation or pelvic nephritis
 \item $^\star$\textbf{PediaLanguages} \citep{morsey2012dbpedia} involves the semantic web of official languages spoken by people living in different countries
 \item $^\star$\textbf{Bottlenose Dolphins}\footnote{Publicly available: \url{http://www-personal.umich.edu/~mejn/netdata/}} \citep{lusseau2003bottlenose} describes a network of frequent community associations of $62$ dolphins living in Doubtful Sound, New Zealand.
\end{itemize}
Their brief descriptions are shown in Table~\ref{tab1}.
\begin{table}[!htbp]
\caption{A description of tested datasets where $|\mathcal{G}|$ (resp. $|\mathcal{M}|$) is the number of objects (resp. attributes) while $|L|$ and $n$ are the lattice size and the number of shared concepts between $\mathbb{K}_R$  and $\mathbb{K}_T$, respectively.}
\label{tab1} 
\centering
\begin{tabular}{p{2.8cm}p{1.5cm}p{1.5cm}p{1.9cm}p{1.5cm}}
\noalign{\smallskip} \hline\noalign{\smallskip}
Dataset ($\mathbb{K}$) & $|\mathcal{G}|$ & $|\mathcal{M}|$ &  $|L|$ & $n$    \\
\noalign{\smallskip} \hline\noalign{\smallskip}
$^\ddagger$Dirichlet & 2000   & 15 & 18,166 &  2903\\
\noalign{\smallskip} \noalign{\smallskip}
$^\ddagger$Coin-Toss & 793   & 10 & 913 &  645 \\
\noalign{\smallskip} \noalign{\smallskip}
$^\star$LinkedIN & 1269   & 34 & 4847 &   1473 \\
\noalign{\smallskip} \noalign{\smallskip}
$^\star$Diagnosis & 120 & 17 & 88 & 81  \\
\noalign{\smallskip} \noalign{\smallskip}
$^\star$PediaLanguages & 316 & 169 & 188 &  22  \\
\noalign{\smallskip} \noalign{\smallskip}
$^\star$Dolphins & 62 & 62 & 282 & 16 \\
\noalign{\smallskip}
\noalign{\smallskip} \hline\noalign{\smallskip}
\toprule\noalign{\smallskip}
\end{tabular}
\end{table}
Subsequently, we compared the results of $\mathcal{CR}(c)$ using \textit{the arithmetic average} as the activation $\mathcal{F}$ against the \textbf{intentional Stability} \citep{Kuznetsov2018}, which is currently the state-of-the-art interestingness index. We then consider the traditional approach \citep{buzmakov2014concept} to validate the two relevance measures. That is, we apply the following scheme: 
\begin{enumerate}
    \item Divide the dataset $\mathbb{K} = (\mathcal{G},\mathcal{M},\mathcal{I})$ horizontally into two disjoint subsets $\mathbb{K}_R = (\mathcal{G}_1,\mathcal{M},\mathcal{I}_1)$ and $\mathbb{K}_T = (\mathcal{G}_2,\mathcal{M},\mathcal{I}_2)$ such that $\mathcal{G}= \mathcal{G}_1 \cup \mathcal{G}_2$. 
    \item Extract the two sets $\mathcal{S}_r$ and $\mathcal{S}_t$ of the shared formal concepts from $\mathbb{K}_R$ and $\mathbb{K}_T$, respectively. Note that a concept $c_r=(A_r,B_r) \in \mathcal{S}_r$ and its corresponding one $c_t=(A_t,B_t) \in \mathcal{S}_t$ are shared if they have the same intent but not necessarily the same extent. 
    \item Use $\mathbb{K}_R$ as a reference dataset to calculate the underlying relevance index (e.g., Conceptual Relevance and stability) of the shared concepts in $\mathcal{S}_r$ while using $\mathbb{K}_T$ as a test dataset to evaluate the relevance index values of the corresponding shared concepts in $\mathcal{S}_t$. It is obvious that $n=|\mathcal{S}_r|=|\mathcal{S}_t|$. 
    \item Record the score list $\{(x_i,y_i)\}_{i=1}^{n}$, where $x_i$ and $y_i$ are the relevant measures of the $i$-th concept in $\mathcal{S}_r$ and its corresponding concept in $\mathcal{S}_t$, respectively 
    \item Draw each pair $(x_i,y_i)$ as a point in a 2D-plot so that the best case is $y_i=x_i$. This means that the tested relevance index produces correct results if there is a strong linear relationship between the relevant evaluations $\{x_i\}_{i=1}^{|\mathcal{S}_r|}$ and $\{y_i\}_{i=1}^{|\mathcal{S}_t|}$. We therefore consider the underlying interestingness measure to be accurate if its relevance values for the shared formal concepts of the reference set $\mathcal{S}_r$ are close or equal to the relevance values of the corresponding formal concepts obtained from the test set $\mathcal{S}_t$. 
\end{enumerate}

Based on this scheme, we consider the following two metrics to assess the accuracy (i.e. the strength of its linearity relationship) and the performance of the results: 

1) The Pearson correlation coefficient $\xi$ (and its scatter plot): 
\begin{equation}\label{pearson}
    \xi =  \frac{\sum_{i=1}^{n} x_i y_i - n \Bar{x}\Bar{y}}{\sqrt{(\sum_{i=1}^{n} x_i^2 - n \Bar{x}^2)} \sqrt{(\sum_{i=1}^{n} y_i^2 - n \Bar{y}^2)}}  
\end{equation}
Where $\Bar{x} = \frac{1}{n} \sum_{i=1}^{n} x_i$ and $\Bar{y} = \frac{1}{n} \sum_{i=1}^{n} y_i$ are the mean values of $\{x_i\}_{i=1}^{|\mathcal{S}_r|}$ and $\{y_i\}_{i=1}^{|\mathcal{S}_t|}$ respectively. We recall that $n = |\mathcal{S}_r|=|\mathcal{S}_t|$ is the number of the shared concepts.

2) The average elapsed time $\tau$:
\begin{equation}\label{Timeeq}
    \tau =  \frac{1}{2}\big[\frac{\sum_{c_i \in  \mathcal{S}_r} t_i}{|\mathcal{S}_r|} + \frac{\sum_{c_j \in  \mathcal{S}_t} t_j}{|\mathcal{S}_t|} \big]
\end{equation}
Where $t_{i}$ and $t_{j}$ are the elapsed times for computing the underlying index of the concept $c_i \in \mathcal{S}_r$ and its corresponding one $c_j \in \mathcal{S}_t$ respectively.

All the experiments were run on an Intel(R) Core-i7 CPU @2.6GHz computer with 16 GB of memory under macOS Mojave. We implemented the two relevance indices as an extension to the Python package called \emph{Concepts 0.7.11}
, which is implemented by Sebastian Bank\footnote{Publicly available: \url{https://pypi.python.org/pypi/concepts}}.

\subsection{Results}\label{resultsss}
We conduct our experimental evaluations through two experiments.\\

\noindent
\textbf{Experiment I.}
The first experiment is dedicated to answering Q1. In line with the scheme explained above, we first divide each one of the six underlying datasets into reference $\mathbb{K}_R$ and tested $\mathbb{K}_T$ subsets. Two relevance indices, namely \textbf{Conceptual Relevance and stability}, are then computed on the extracted sets of shared concepts. On that basis, we calculate their Pearson correlation coefficients using their recorded score lists. 

\begin{figure}[!htbp]
  \begin{center}
          \includegraphics[width=45mm]{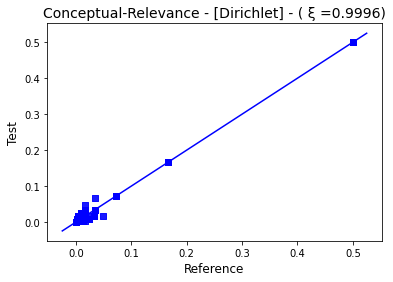}  
          \includegraphics[width=45mm]{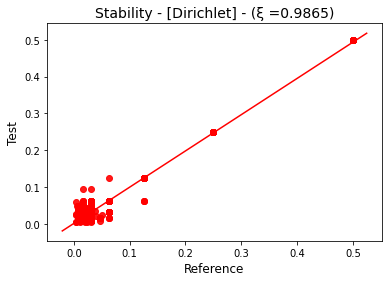} \\
      \includegraphics[width=45mm]{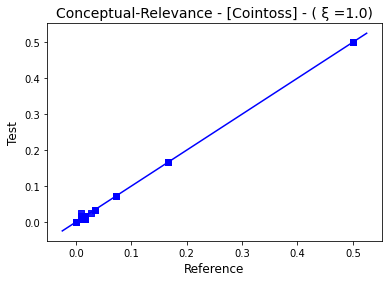}  
          \includegraphics[width=45mm]{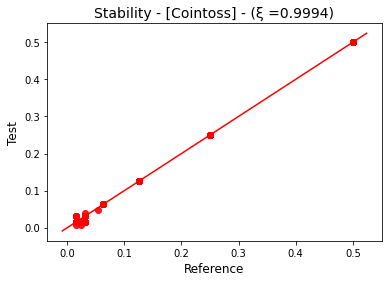} \\
          \includegraphics[width=45mm]{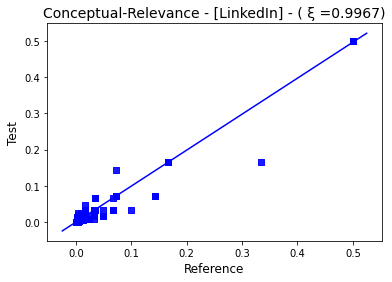}  
          \includegraphics[width=45mm]{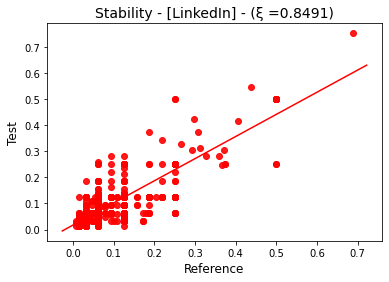} \\
          \includegraphics[width=45mm]{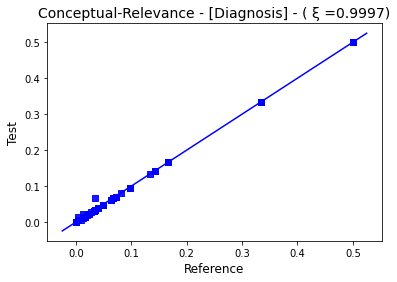}  
          \includegraphics[width=45mm]{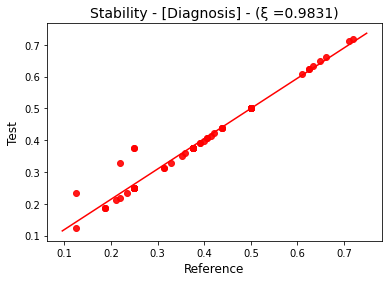} \\
           \includegraphics[width=47mm]{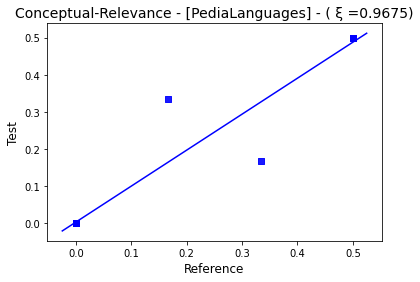}  
          \includegraphics[width=45mm]{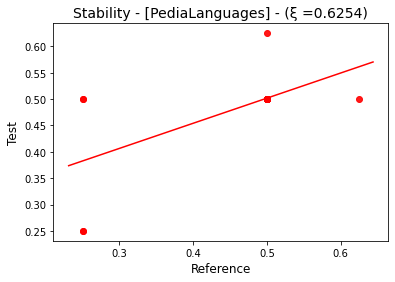} \\
          \includegraphics[width=45mm]{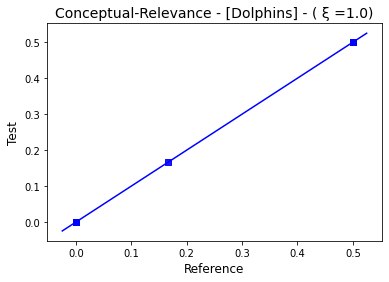}  
          \includegraphics[width=45mm]{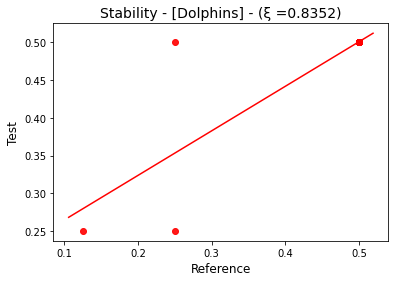} \\
    \end{center}
      \caption{The Pearson correlation scatter plot of reference $\mathbb{K}_R$ vs test $\mathbb{K}_T$ of the two relevancy measures: (\textit{left column}) Conceptual relevance (arithmetic), and (\textit{right column}) Stability on the shared concepts of the six tested datasets. The Pearson correlation coefficient $\eta$ appears between parentheses.}
      \label{Exp1}
\end{figure}

Figure~\ref{Exp1} displays the Pearson correlation scatter plot of reference $\mathbb{K}_R$ vs test $\mathbb{K}_T$ on the shared concepts. Notation used to label each figure is as follows: \textit{name of the index - [name of the dataset] - (Pearson correlation coefficient value $\xi$ calculated as in Eq.~\ref{pearson})}. Overall, Conceptual Relevance is the more accurate of the two compared indices, achieving the best Pearson correlation coefficients on the six datasets. Stability comes close behind the Conceptual Relevance on Dirichlet, CoinToss and Diagnosis datasets, but considerably further behind with large margins otherwise. $\mathcal{CR}(c)$ is at least $14 \%$ more accurate than stability on LinkedIn dataset, $33 \%$ more accurate than it on the Pedia-Languages dataset, and $16 \%$ more accurate than it on the Dolphins dataset.  

\noindent
\textbf{Experiment II.}
\begin{figure}[!htbp]
  \begin{center}
      \includegraphics[width=50mm]{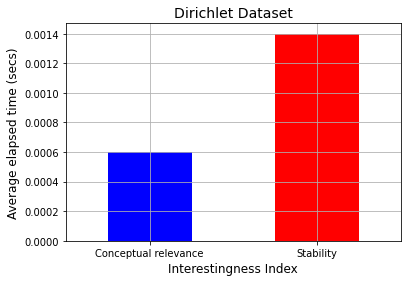} 
      \includegraphics[width=50mm]{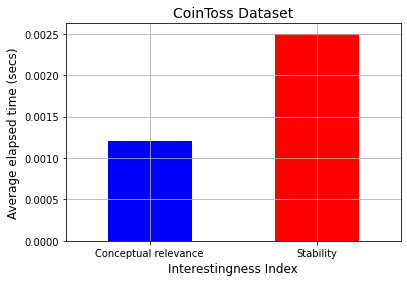} \\
      \includegraphics[width=50mm]{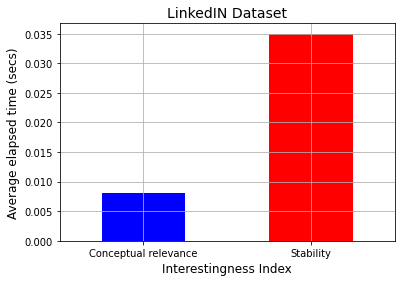} 
      \includegraphics[width=50mm]{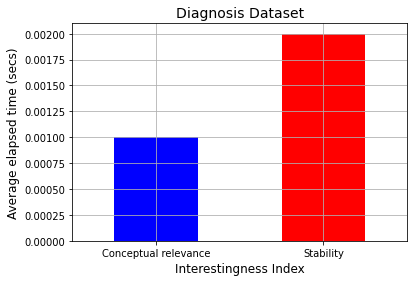} \\
      \includegraphics[width=50mm]{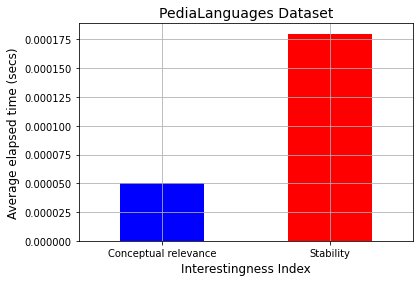}
      \includegraphics[width=50mm]{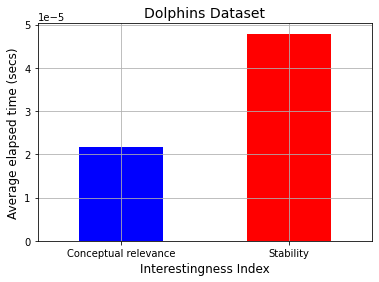}
    \end{center}
      \caption{The average elapsed time $\tau$ of the two relevancy measures: Conceptual relevance (arithmetic), and stability on the shared concepts of the six underlying datasets.} 
      \label{time}
\end{figure}
This experiment is performed to answer Q2. We are interested here in assessing the performance of the indices. That is, we rerun Experiment I while reporting their computational times as in Eq.~\ref{Timeeq}. Figure~\ref{time} shows the average elapsed time $\tau$ of the two relevance indices, namely the $\mathcal{CR}(c)$ and stability, on the shared concepts of the six underlying datasets. $\mathcal{CR}(c)$ dominates stability on all datasets tested. It performs four times faster than the stability on the LinkedIn dataset, three times faster on the Pedia-Languages and at least twice as fast as stability on the Dirichlet, CoinToss, Diagnosis and Dolphins datasets. 
\subsection{Discussion}
In terms of accuracy, the results of Experiment I in Subsection~\ref{resultsss} suggest that the Conceptual relevance index outperforms the stability index. It improves the assessment of the concept quality in two ways. First, rather than relying on generators as in stability which may result in redundant patterns, the Conceptual relevance precisely quantifies the amount of both actionable minimal generators and relevant attributes in the concept intent, with the ability to encapsulate that in a single measurement and in a variety of ways using potential activation functions. Second, and unlike the stability index, $\mathcal{CR}(c)$ produces robust relevance scores to effectively discriminate between small and neighbour concepts (i.e., those with an immediate link). This is due to the fact that neighbor concepts cannot have identical set of minimal generators, and hence may have distinct $\mathcal{CR}(c)$ values.      

The results of Experiment II about the performance of the two indices in Subsection~\ref{resultsss} show that the arithmetic $\mathcal{CR}(c)$ considerably prevails over stability. This is due to the fact that it addresses the limitation of the stability, which presents the threat of  exponential time and space complexity to validate the probability of satisfying the derivation condition of the concept intent. This is attributable to the virtue of counting the number of relevant attributes in the intent instead of subsets in the intent power set, and also leveraging faces w.r.t. the upper covers to efficiently identify the set of minimal generators. From a computational perspective, $\mathcal{CR}(c)$ requires polynomial time and space complexity that does not depend on the size of the intent power set.

\section{Conclusion}\label{Con}
Our work here targeted the vital challenge of extracting interesting concepts from a concept lattice using a new relevance score. Since the comparative analysis of the local conceptual structures of the lattice often helps discover new knowledge, we believe that there is a clear gap in the existing FCA literature on how to identify important substructures in a concept intent such as its relevant attributes and generators. On that basis, we proposed Conceptual Relevance index, a scalable interestingness index to assess the quality of the concept. The novelty of the $\mathcal{CR}$ index is twofold: (i) first, it contrasts the attributes of a concept intent to mine the relevant ones that maintain its conceptual structure, and (ii) second, it leverages the strength of minimal generators to quantify the most relevant local substructures of the intent, taking advantage of the fact that minimal generators frequently lead to more relevant (with less redundancy) patterns than non minimal ones. $\mathcal{CR}$ requires only a polynomial time and space complexity in the number of the concept upper covers and minimal generators. In addition, its formula is flexible and can be easily adjusted with several variants of its activation function. The thorough empirical study on several synthetic and real-life concept lattices 
(see Section \ref{Exp}) 
shows that the $\mathcal{CR}$ can assess the quality of concepts in a more accurate and efficient manner than state-of-the-art relevance indices like stability. We are presently conducting additional empirical tests on larger datasets to confirm the findings.

\bibliographystyle{unsrtnat}
\bibliography{References}  

\begin{thebibliography}{22}
\providecommand{\natexlab}[1]{#1}
\providecommand{\url}[1]{\texttt{#1}}
\expandafter\ifx\csname urlstyle\endcsname\relax
  \providecommand{\doi}[1]{doi: #1}\else
  \providecommand{\doi}{doi: \begingroup \urlstyle{rm}\Url}\fi

\bibitem[Ganter and Wille(1999)]{Ganter+1999}
Bernhard Ganter and Rudolf Wille.
\newblock \emph{Formal Concept Analysis: Mathematical Foundations}.
\newblock Springer-Verlag New York, Inc., 1999.
\newblock Translator-C. Franzke.

\bibitem[Belohl{\'{a}}vek and Macko(2011)]{belohlavek2011selecting}
Radim Belohl{\'{a}}vek and Juraj Macko.
\newblock Selecting important concepts using weights.
\newblock In \emph{ICFCA}, pages 65--80. Springer, 2011.

\bibitem[Kuznetsov and Makhalova(2018)]{Kuznetsov2018}
Sergei~O Kuznetsov and Tatiana Makhalova.
\newblock On interestingness measures of formal concepts.
\newblock \emph{Information Sciences}, 442:\penalty0 202--219, 2018.

\bibitem[Belohl{\'{a}}vek and Trnecka(2013)]{Belohlavek2013}
Radim Belohl{\'{a}}vek and Martin Trnecka.
\newblock Basic level in formal concept analysis: Interesting concepts and
  psychological ramifications.
\newblock In \emph{{IJCAI} 2013, Proceedings of the 23rd International Joint
  Conference on Artificial Intelligence, Beijing, China, August 3-9, 2013},
  pages 1233--1239, 2013.

\bibitem[Torim and Lindroos(2008)]{torim2008sorting}
Ants Torim and Karin Lindroos.
\newblock Sorting concepts by priority using the theory of monotone systems.
\newblock In \emph{International Conference on Conceptual Structures}, pages
  175--188. Springer, 2008.

\bibitem[Cheng et~al.(2006)Cheng, Ke, and Ng]{cheng2006delta}
James Cheng, Yiping Ke, and Wilfred Ng.
\newblock $\backslash$delta-tolerance closed frequent itemsets.
\newblock In \emph{Sixth International Conference on Data Mining (ICDM'06)},
  pages 139--148. IEEE, 2006.

\bibitem[Moerchen et~al.(2011)Moerchen, Thies, and
  Ultsch]{moerchen2011efficient}
Fabian Moerchen, Michael Thies, and Alfred Ultsch.
\newblock Efficient mining of all margin-closed itemsets with applications in
  temporal knowledge discovery and classification by compression.
\newblock \emph{Knowledge and Information Systems}, 29\penalty0 (1):\penalty0
  55--80, 2011.

\bibitem[Belohlavek and Trnecka(2013)]{belohlavek2013basic}
Radim Belohlavek and Martin Trnecka.
\newblock Basic level in formal concept analysis: Interesting concepts and
  psychological ramifications.
\newblock In \emph{Twenty-Third International Joint Conference on Artificial
  Intelligence}, 2013.

\bibitem[Kuznetsov et~al.(2007)Kuznetsov, Obiedkov, and
  Roth]{kuznetsov2007reducing}
Sergei Kuznetsov, Sergei Obiedkov, and Camille Roth.
\newblock Reducing the representation complexity of lattice-based taxonomies.
\newblock In \emph{International Conference on Conceptual Structures}, pages
  241--254. Springer, 2007.

\bibitem[Roth et~al.(2008)Roth, Obiedkov, and Kourie]{roth2008succinct}
Camille Roth, Sergei Obiedkov, and Derrick~G Kourie.
\newblock On succinct representation of knowledge community taxonomies with
  formal concept analysis.
\newblock \emph{International Journal of Foundations of Computer Science},
  19\penalty0 (02):\penalty0 383--404, 2008.

\bibitem[Zhi(2014)]{zhi2014calculation}
Hui-lai Zhi.
\newblock On the calculation of formal concept stability.
\newblock \emph{Journal of Applied Mathematics}, 2014, 2014.

\bibitem[Pfaltz and Taylor(2002)]{pfaltz2002closed}
John~L Pfaltz and Christopher~M Taylor.
\newblock Closed set mining of biological data.
\newblock In \emph{BIOKDD}, pages 43--48, 2002.

\bibitem[Bastide et~al.(2000)Bastide, Pasquier, Taouil, Stumme, and
  Lakhal]{bastide2000mining}
Yves Bastide, Nicolas Pasquier, Rafik Taouil, Gerd Stumme, and Lotfi Lakhal.
\newblock Mining minimal non-redundant association rules using frequent closed
  itemsets.
\newblock In \emph{International Conference on Computational Logic}, pages
  972--986. Springer, 2000.

\bibitem[Ganter and Wille(2012)]{ganter2012formal}
Bernhard Ganter and Rudolf Wille.
\newblock \emph{Formal concept analysis: mathematical foundations}.
\newblock Springer Science \& Business Media, 2012.

\bibitem[Szathmary et~al.(2014)Szathmary, Valtchev, Napoli, Godin, Boc, and
  Makarenkov]{szathmary2014fast}
Laszlo Szathmary, Petko Valtchev, Amedeo Napoli, Robert Godin, Alix Boc, and
  Vladimir Makarenkov.
\newblock A fast compound algorithm for mining generators, closed itemsets, and
  computing links between equivalence classes.
\newblock \emph{Annals of Mathematics and Artificial Intelligence}, 70\penalty0
  (1):\penalty0 81--105, 2014.

\bibitem[Felde and Hanika(2019)]{felde2019formal}
Maximilian Felde and Tom Hanika.
\newblock Formal context generation using dirichlet distributions.
\newblock In \emph{International Conference on Conceptual Structures}, pages
  57--71. Springer, 2019.

\bibitem[Felde et~al.(2020)Felde, Hanika, and Stumme]{felde2020null}
Maximilian Felde, Tom Hanika, and Gerd Stumme.
\newblock Null models for formal contexts.
\newblock \emph{Information}, 11\penalty0 (3):\penalty0 135, 2020.

\bibitem[Rossi and Ahmed(2015)]{nr}
Ryan~A. Rossi and Nesreen~K. Ahmed.
\newblock The network data repository with interactive graph analytics and
  visualization.
\newblock In \emph{AAAI}, 2015.
\newblock URL \url{http://networkrepository.com}.

\bibitem[Czerniak and Zarzycki(2003)]{czerniak2003application}
Jacek Czerniak and Hubert Zarzycki.
\newblock Application of rough sets in the presumptive diagnosis of urinary
  system diseases.
\newblock In \emph{Artificial intelligence and security in computing systems},
  pages 41--51. Springer, 2003.

\bibitem[Morsey et~al.(2012)Morsey, Lehmann, Auer, Stadler, and
  Hellmann]{morsey2012dbpedia}
Mohamed Morsey, Jens Lehmann, S{\"o}ren Auer, Claus Stadler, and Sebastian
  Hellmann.
\newblock Dbpedia and the live extraction of structured data from wikipedia.
\newblock \emph{Program}, 46\penalty0 (2):\penalty0 157--181, 2012.

\bibitem[Lusseau et~al.(2003)Lusseau, Schneider, Boisseau, Haase, Slooten, and
  Dawson]{lusseau2003bottlenose}
David Lusseau, Karsten Schneider, Oliver~J Boisseau, Patti Haase, Elisabeth
  Slooten, and Steve~M Dawson.
\newblock The bottlenose dolphin community of doubtful sound features a large
  proportion of long-lasting associations.
\newblock \emph{Behavioral Ecology and Sociobiology}, 54\penalty0 (4):\penalty0
  396--405, 2003.

\bibitem[Buzmakov et~al.(2014)Buzmakov, Kuznetsov, and
  Napoli]{buzmakov2014concept}
Aleksey Buzmakov, Sergei~O. Kuznetsov, and Amedeo Napoli.
\newblock Is concept stability a measure for pattern selection?
\newblock In \emph{Proceedings of the Second International Conference on
  Information Technology and Quantitative Management, {ITQM} 2014, National
  Research University Higher School of Economics (HSE), Moscow, Russia, June
  3-5, 2014}, pages 918--927, 2014.

\end{thebibliography}

\end{document}